\documentclass[runningheads]{llncs}

\usepackage{eccv}

\usepackage{eccvabbrv}
\usepackage{pifont}
\usepackage{graphicx}
\usepackage{booktabs}
\usepackage{multirow}
\usepackage{threeparttable}
\usepackage{bm}
\usepackage{bbding}
\usepackage{bbm}
\usepackage[accsupp]{axessibility}  

\usepackage[pagebackref,breaklinks,colorlinks,citecolor=eccvblue]{hyperref}

\usepackage{orcidlink}

\begin{document}

\title{EPL: Evidential Prototype Learning for Semi-supervised Medical Image Segmentation} 


\author{Yuanpeng He\inst{1,2}}

\institute{Key Laboratory of High Confidence Software Technologies (Peking University), Ministry of Education, Beijing, 100871, China \and
School of Computer Science, Peking University, Beijing, 100871, China
\email{heyuanpeng@stu.pku.edu.cn}\\
}

\maketitle

\begin{abstract}
  Although current semi-supervised medical segmentation methods can achieve decent performance, they are still affected by the uncertainty in unlabeled data and model predictions, and there is currently a lack of effective strategies that can explore the uncertain aspects of both simultaneously. To address the aforementioned issues, we propose Evidential Prototype Learning (EPL), which utilizes an extended probabilistic framework to effectively fuse voxel probability predictions from different sources and achieves prototype fusion utilization of labeled and unlabeled data under a generalized evidential framework, leveraging voxel-level dual uncertainty masking. The uncertainty not only enables the model to self-correct predictions but also improves the guided learning process with pseudo-labels and is able to feed back into the construction of hidden features. The method proposed in this paper has been experimented on LA, Pancreas-CT and TBAD datasets, achieving the state-of-the-art performance in three different labeled ratios, which strongly demonstrates the effectiveness of our strategy.
  \keywords{Semi-supervised medical segmentation \and Evidential prototype learning \and Dual uncertainty masking}
\end{abstract}

\section{Introduction}
Medical image segmentation (MIS) has received extensive attention from researchers with the wide application of Computer-aided detection and diagnosis (CAD), which assists doctors in better browsing medical images and making more accurate diagnoses and decisions. Compared with natural image segmentation, the low contrast and significant object shape variation as well as this kind of features of medical images make them more challenging to segment. Recently, researchers have applied deep learning algorithms like convolution neural network (CNN) which is capable of representing and learning features. It is noteworthy that U-Net \cite{DBLP:conf/miccai/RonnebergerFB15}, CA-Net \cite{DBLP:journals/tmi/GuWSHADOVZ21} and a series of encoder-decoder networks contribute to improving the segmentation accuracy on medical images under the fully-supervised setting. However, a large number of training images and their fine-grained annotations which are extremely expensive and labor-intensive, are indispensable to achieve such great performance, which severely impedes the application of image segmentation in real scenarios.

To obtain appreciable segmentation results in the experimental environment with limited annotated data, researchers propose numerous semi-supervised learning-based (SSL) approaches which also have made great progress in segmentation performance. These kinds of methods primarily aim to explore how to sufficiently exploit the intrinsic valuable feature information contained in abundant unlabeled data that is relatively uncomplicated to acquire, to guide the learning process of the model under the supervision of a small percentage of labeled data. In general, there are numerous techniques utilized in SSL approaches such as consistency regularization, pseudo-labeling and uncertainty estimation. As for methods based on pseudo-labeling like co-training methods \cite{DBLP:journals/mia/WangPPZZD21, DBLP:journals/pr/PengEPD20, 9093608} and self-training methods \cite{DBLP:conf/cvpr/KwonK22, DBLP:conf/miccai/BaiOSSRTGKMR17}, their learning paradigm is that employing a pre-trained network to generate pseudo-labels for unlabeled data and then utilize some high-quality pseudo-labels to promote model learning in reverse. However, the low quality of pseudo-labels may seriously affect the segmentation performance because they are not capable of providing model with beneficial supervision similar to labeled data. Some researchers focus on reducing the interference of irrelevant unlabeled data and put forward some uncertainty estimation methods like Monte Carlo dropout \cite{DBLP:conf/miccai/YuWLFH19} and ensemble-based methods \cite{shi2021inconsistency}. Moreover, some advanced metrics like information entropy which is widely used, are utilized to measure the level of uncertainty. They set a threshold to judge which unlabeled data segmentation masks can be used as pseudo labels during the training phase. Besides, regarding consistency regularization-based methods, all of them follow a smoothness hypothesis that the labels of the data points that are near to one another in the latent space should be comparable or the same. Such a pattern guides the model to generate a consistent segmentation mask even though the input is processed by some transformations, such as data augmentation and some perturbations. All data will be consistently processed using a consistency loss function, like L2 Loss when it comes to prediction optimization. Additionally, contrastive learning has also gradually become a prevalent technique in the field of medical image segmentation because of its success applied in self-supervised learning. For instance, to improve the feature representation capability, auxiliary class-level \cite{LIU2022102092} or image-level \cite{DBLP:journals/mia/WangZZWZZW22} constraints are devised.

Despite recent advancements, the performance of pseudo-labeling-based methods still faces challenges in designing a suitable reliable threshold that can be utilized in multiple different tasks, which is limited by the innate intricacy of uncertainty. Moreover, the assumption that most consistency constraints follow requires that decision boundary only exists in low-density regions in latent space, which may not properly guide model learning because the valuable feature information contained in unlabeled data is not employed fully. Regarding these problems, researchers introduce prototype alignment into the field of SSL. Prototype-based approaches aim at exploiting the structure information in both labeled and unlabeled data and improving the feature embeddings' distribution among different categories, which may be capable of eliminating the aforementioned problems. However, current methods based on prototype learning always utilize voxel-averaging to obtain class prototypes which is only suitable for labeled data. They still principally depend on a large percentage of labeled data and are not able to learn a unified prototype for both labeled and unlabeled data. It indicates that there is a tremendous space for improvement in accurately depicting the embedding space's distribution.

To solve the aforementioned problems, we propose the framework of evidential prototype learning in this paper. Firstly, we extend the probabilistic framework by introducing multi-objective sets into evidential deep learning, allowing for more detailed modeling of probability distributions. Secondly, to effectively fuse the predictions of evidential multi-classifiers, we propose the use of Dempster's combination rule to strengthen the predictions for the same voxel location from different sources further. It is worth noting that under the generalized evidential deep learning framework, the uncertainty corresponding to the evidence participates in the fusion process, achieving a more refined allocation of credibility. Thirdly, considering the effectiveness of belief entropy in measuring evidence uncertainty, we design a dual uncertainty measurement mechanism that combines the existing uncertainty measure and belief entropy to measure the uncertainty of each voxel's prediction. Specifically, on the one hand, we allow the student model to self-correct its learning based on the uncertainty of its predictions in the learning process from labeled data; on the other hand, we use the uncertainty contained in pseudo-labels to guide the student model to learn on unlabeled data, rather than using the model's inherent uncertainty. Fourth, we have also redesigned the optimization function of evidential deep learning, aiming for the model's predictions to remain as consistent with the labels while avoiding forced optimization with objects of high uncertainty, which could exacerbate prediction biases in remaining parts. Overall, the uncertainty-generalized evidential learning pattern not only considers the model's circumstances but also fully accounts for the uncertainty contained in unlabeled data. Fifth, in generating prototypes for labeled and unlabeled data, we utilize the uncertainty produced in the generalized evidential learning framework to mask model output features, reducing the impact of unreliable features on prototype generation. In summary, the contributions of the proposed method can be given as:

\textbf{1.} The evidential prototype learning framework extends the probabilistic framework by incorporating multi-objective sets into evidential deep learning for more refined probability distributions. It employs Dempster’s combination rule for fusing evidential multi-classifier predictions, integrates belief entropy for dual uncertainty measurement, and guides learning through uncertainty in labeled and unlabeled data, thereby improving prediction accuracy and credibility allocation.

\textbf{2.} The framework redesigns the optimization function to avoid biases by not forcing optimization with high-uncertainty objects and utilizes generated uncertainties to mask unreliable features in prototype generation for both labeled and unlabeled data, enhancing the model’s ability to deal with inherent uncertainties and improving the reliability of its predictions.

\textbf{3.} The method proposed in this paper achieves state-of-the-art performance on a majority of metrics across three annotation ratios in the Left Atrium (LA), Pancreas-CT, and Type B Aortic Dissection (TBAD) datasets. Notably, the proposed method significantly outperforms existing methods on the TBAD dataset, achieving superior performance with only 5\% of labeled data compared to other methods that utilize 20\% of labeled data.

\section{Related Work}
\subsection{Semi-supervised Medical Image Segmentation}
In this section, some relevant semi-supervised medical image segmentation works that aim to fully exploit the valuable information from unlabeled data will be introduced briefly. The existing methods can be roughly divided into four categories: pseudo-labeling based methods \cite{DBLP:conf/cvpr/KwonK22, DBLP:conf/miccai/BaiOSSRTGKMR17, DBLP:conf/ijcnn/ArazoOAOM20, article, DBLP:conf/nips/ZhangWHWWOS21}, consistency regularization based methods \cite{DBLP:conf/nips/BachmanAP14, DBLP:conf/nips/SajjadiJT16, DBLP:conf/miccai/BortsovaDHKB19, DBLP:conf/icml/XuSYQLSLJ21,  DBLP:conf/aaai/LuoCSW21, DBLP:conf/miccai/LuoLCSCZCWZ21}, uncertainty-aware methods \cite{9093608, DBLP:conf/miccai/YuWLFH19, DBLP:conf/miccai/WangZTZSZH20}, prototype-based methods \cite{DBLP:journals/pami/WuFHHMZ23, DBLP:conf/cvpr/WangWSFLJWZL22, DBLP:journals/pami/WuFHHMZ23, DBLP:journals/titb/XuWLYY000T22}. Pseudo-labeling-based methods can also be classified into three common methods including self-training methods \cite{DBLP:conf/cvpr/KwonK22, DBLP:conf/miccai/BaiOSSRTGKMR17}, co-training methods \cite{DBLP:journals/mia/WangPPZZD21, DBLP:journals/pr/PengEPD20, 9093608} and adversarial learning methods \cite{DBLP:conf/miccai/LiZH20, DBLP:conf/miccai/FangL20, DBLP:conf/miccai/ZhangYCFHC17}. For instance, Bai et al. \cite{DBLP:conf/miccai/BaiOSSRTGKMR17} proposes a self-training framework that trains a segmentation network with both labeled and unlabeled data whose parameters and predictions for unlabeled data will be updated alternately. Xia et al. \cite{9093608} utilize different views of 3D data to achieve co-training and propose an uncertainty-weighted label fusion mechanism based on Bayesian deep learning to generate higher-quality pseudo-labels. Zhang et al. \cite{DBLP:conf/nips/ZhangWHWWOS21} proposes a tactic called Curriculum Pseudo Labeling (CPL) to adaptively adjust the threshold of pseudo labels during the training process. Zhang et al. \cite{DBLP:conf/miccai/ZhangYCFHC17} introduces an evaluation network (EN) to encourage to generate unanimous segmentation for labeled and unlabeled data, and proposes a segmentation network (SN) to produce segmentation masks for unlabeled data that are similar to the segmentation results of annotated data. Such a learning strategy can guide the SN to generate outputs expected by the EN. However, due to the complexity and instability of model training, the performance of the above methods is limited. 

Consistency regularization-based methods have attracted more attention from researchers to be applied in semi-supervised MIS. They aim to exploit unlabeled data to produce consistent outputs even though some perturbations are added to inputs. Bortsova et al. \cite{DBLP:conf/miccai/BortsovaDHKB19} proposes an innovative paradigm that encourages the model to output consistent segmentation masks when both labeled and unlabeled data are processed by a series of transformations. Luo et al. \cite{DBLP:conf/aaai/LuoCSW21} introduces a dual-task consistency regularization to enforce consistency between the segmentation maps derived from the level set representations and the directly predicted segmentation maps. This regularization is applied to both labeled and unlabeled data, enabling the model to leverage unlabeled data effectively to improve segmentation performance. Ouali et al. \cite{DBLP:conf/cvpr/OualiHT20} proposes a framework based on cross-consistency training (CCT) to enforce consistency between the predictions of the main decoder and those of auxiliary decoders which work on different perturbed versions of the encoder's output, enhancing the robustness and accuracy of the segmentation model. Except for designing different perturbations, researchers introduce uncertainty estimation to further improve the performance of segmentation. Yu et al. \cite{DBLP:conf/miccai/YuWLFH19} proposes an uncertainty-aware strategy to exploit uncertainty information that is estimated through Monte Carlo sampling so that the student model is encouraged to produce consistent predictions under various perturbations by focusing on area with low uncertainty. Wang et al. \cite{DBLP:conf/miccai/WangZTZSZH20} introduces a double-uncertainty weighted method into a teacher-student model to extend segmentation uncertainty estimation to feature uncertainty, which enables the model to capture information among channels, revealing the capability to handle complex relationships within the data.

Different from the aforementioned methods, prototype-based methods aim to construct prototypes that are regarded as class representatives and can delegate essential characteristics of different anatomical structures or pathological regions for both labeled and unlabeled data, which enables the model to leverage valuable information contained in unlabeled images with prototypes. For instance, Wang et al. \cite{DBLP:conf/cvpr/WangWSFLJWZL22} consider the importance of all pixels and separate reliable and unreliable pixels with a threshold by calculating the prediction's entropy. Then, by adaptively adjusting the threshold of differentiation, the prediction becomes increasingly accurate. Wu et al. \cite{DBLP:journals/pami/WuFHHMZ23} devises a strategy named Cross-Image Semantic Consistency guided Rectifying (CISC-R) for semi-supervised semantic segmentation. It refines pseudo labels using pixel-level correspondence within the same class by querying a guiding labeled image and creating a reliable CISC map for rectification.

\subsection{Evidential Deep Learning and Evidence Theory}
Evidence Theory, also known as Dempster-Shafer Theory (DST) \cite{DBLP:series/sfsc/Dempster08a,DBLP:journals/ijar/Shafer16}, is a mathematical framework for modeling epistemic uncertainty. It provides a flexible tool for combining evidence from different sources and making decisions based on incomplete and uncertain information. Unlike traditional probability theory, which requires precise probabilities for each event, DST works with belief functions that offer a range of probability values, allowing for a more nuanced expression of uncertainty. As deep learning technology continues to evolve, ensuring the reliability of deep learning models and reducing the uncertainty of prediction results have increasingly become hot topics of research. Drawing from the principles of Dempster-Shafer theory of evidence and Subjective Logic \cite{DBLP:books/sp/Josang16}, evidential deep learning (EDL) \cite{amini2020deep,DBLP:conf/nips/SensoyKK18} is introduced to overcome the limitations of softmax-based classifiers discussed. Instead of estimating class probabilities directly, EDL first gathers evidence for each class, and then constructs a Dirichlet distribution of class probabilities based on the evidence obtained. This approach allows for the quantification of predictive uncertainty through subjective logic. Evidence is conceptualized as the degree of support amassed from the data for classifying a sample into a specific class \cite{DBLP:series/sfsc/LiuY08}, serving as a quantifiable measure of class activation intensity.

\section{Method}
\begin{figure*}
	\centering 
	\includegraphics[scale=0.335]{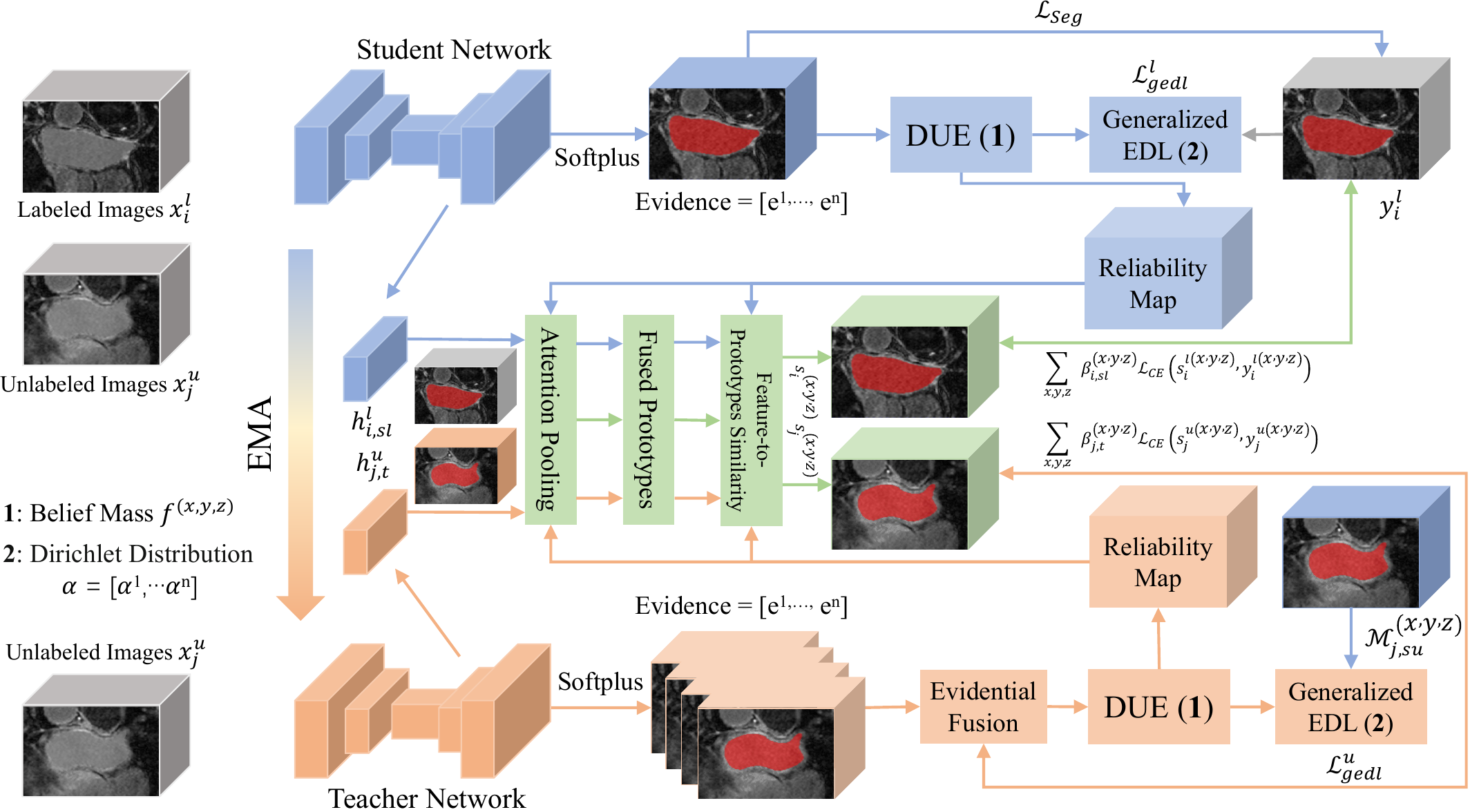}
	\caption{Overview of EPL framework (using Left Atrium dataset for illustration). The labeled and unlabeled images are sent to the student model while the unlabeled images are sent to the teacher model. The output of the student model on the labeled and unlabeled images are supposed to be optimized by utilizing the uncertainty from model prediction and pseudo-labels. The evidential predictions of the teacher model are fused by using the Dempster's combination rule. Besides, the prototype fusion process utilizes masked attention pooling with the generated uncertainty-based reliability map. }
	\label{fig1}
\end{figure*}

Assume there exists a dataset $\mathcal{S} = \{\mathcal{S}^{l}, \mathcal{S}^{u}\}$ in which the labeled set consists of $\mathcal{S}^{l} = \{x_i^l, y_i^l\}_{i = 1}^{L}$ with $L$ samples and unlabeled set consists of $\mathcal{S}^{u} = \{x_j^u\}_{j = N}^{L+M}$ with $M$ samples $(L \ll M)$. Specifically, $x_i^l, x_i^u \in \mathbb{R}^{W \times H \times D}$ denote the input with width $W$, height $H$ and depth $D$ and $y_i^u \in \{0,1,...,N-1\}^{W \times H \times D}$. The proposed framework EPL consists of a teacher-student network and the details of the proposed method are provided in Figure \ref{fig1}.
\subsection{Generalized Probabilistic Framework}
Here, we extend EDL by integrating multiple objective sets in traditional evidence theory \cite{DBLP:series/sfsc/Dempster08a}, implementing evidential fusion-based dual uncertainty evaluation with belief entropy and generalized evidential deep learning. We map the inherent uncertainty of evidence-based deep learning onto the universal set corresponding to the discernment framework of traditional evidence theory. This allows for greater flexibility during the process of evidence-based feature fusion and uncertainty measurement. 

To be specific, let $f_{i,sl,m}^{(x,y,z)}, f_{j,su,m}^{(x,y,z)}, f_{j,t,m}^{(x,y,z)} \in \mathcal{R}^{C}$ denote the belief mass from evidential classifier predictive results \cite{DBLP:conf/nips/SensoyKK18,10.5555/3618408.3618708} for voxel at position $(x,y,z)$ by the segmentation head of student model on $i_{th}$ labeled image, $j_{th}$ unlabeled image and teacher model on $j_{th}$ unlabeled image of the $m_{th}$ classifiers respectively. Besides, $N$ denotes the number of classes. We define basic probability mass assignments for voxel as $\mathcal{M} = \{\{f^{(x,y,z)}(C_n)\}_{n=0}^{N-1}, u^{(x,y,z)}\}$ and $u^{(x,y,z)} = 1 - \sum_{n=0}^{N-1}f^{(x,y,z)}(C_n)$ represent the original uncertainty. Then the generalized probability mass assignments can be given as:
\begin{equation}
    \mathcal{M}^{(x,y,z)} = \{\{f^{(x,y,z)}(C_n)\}_{n=0}^{N-1}, f^{(x,y,z)}(C_N)\},\ C_N = \{C_0,...,C_{N-1}\} 
\end{equation}

\subsection{Evidential Fusion-based Prediction Synthesis}

Different from UPCoL \cite{lu2023upcol}, the teacher model's prediction results are simply an average of the output of multiple classifiers, we utilize Dempsters' combination rule \cite{shafer2016dempster,lee2019dbf} to effectively combine mass assignments from four predictions. We first consider the fusion process of two independent mass assignments $\mathcal{M}^{(x,y,z)}_{j,1}$ and $\mathcal{M}^{(x,y,z)}_{j,2}$ at location $(x,y,z)$ from classifier 1 and 2 for $j_{th}$ unlabeled image:
\begin{equation}
    \mathcal{M}^{(x,y,z)}_{j,1,2} = \mathcal{M}^{(x,y,z)}_{j,1} \oplus \mathcal{M}^{(x,y,z)}_{j,2}
\end{equation}

More specifically, the combination rule can be formulated as (the superscript $(x,y,z)$ is omitted for simplicity):
\begin{equation}\
    f_{j,1,2}(C_n) = \frac{1}{1-\Delta}(f_{j,1}(C_n)f_{j,2}(C_n)+f_{j,1}(C_n)f_{j,2}(C_N) + f_{j,2}(C_n)f_{j,1}(C_N))
\end{equation}
where $\Delta =  \sum_{a \neq b} f_{j,1}(C_a)f_{j,2}(C_b)$ represents the conflicting degree between mass assignments $\mathcal{M}^{(x,y,z)}_{j,1}$ and $\mathcal{M}^{(x,y,z)}_{j,2}$, and $f_{j,1,2}(C_N) = f_{j,1}(C_N)f_{j,2}(C_N)$. $\frac{1}{1-\Delta}$ plays the role of normalization. In general, the fusion process of mass assignments for each voxel from $T$ classifiers can be given as:
\begin{equation}
    \mathcal{M}^{(x,y,z)}_{j,t} = \mathcal{M}^{(x,y,z)}_{j,1} \oplus \mathcal{M}^{(x,y,z)}_{j,2} \oplus ... \oplus \mathcal{M}^{(x,y,z)}_{j,T}
\end{equation}

The fused mass assignments can be used for providing corresponding uncertainty measures and reverted into the form of evidence, which is supposed to be utilized for generation of prototypes and as pseudo-labels for unlabeled images.

\subsection{DUE: Dual Uncertainty Evaluation with Belief Entropy}
With the help of evidential fusion, we can obtain final teacher segmentation results $\mathcal{M}^{(x,y,z)}_{j,t}$ within multiple classifiers. Student model segmentation results for labeled and unlabeled images are denoted by $\mathcal{M}^{(x,y,z)}_{i,sl}$ and $\mathcal{M}^{(x,y,z)}_{j,su}$. Due to the effectiveness of belief entropy \cite{DBLP:journals/chinaf/Deng20}, the uncertainty measure $\mathcal{U}$ synthesizing original uncertainty $u^{(x,y,z)}$ and belief entropy can be given as:
\begin{equation}
    \mathcal{U}(\mathcal{M}^{(x,y,z)}_{v,w}) = - \sum_{n=0}^{N}u_{v,w}^{(x,y,z)} (f_{v,w}^{(x,y,z)}(C_n) log_2 \frac{f_{v,w}^{(x,y,z)}(C_n)}{2^{|C_n|}-1})
\end{equation}
where $v, w = (j,t)\ or \ (i,sl)$ and $|C_n|$ denotes the cardinality of $C_n$ ($|C_n|=1, |C_N|=N$). Then, the voxels with higher uncertainty are considered more ambiguous. We adopt the same policy of UPCoL to generate corresponding reliability map $\beta_{j,t}^{(x,y,z)}$ and $\beta_{i,sl}^{(x,y,z)}$ for each voxel importance assessment. 


\subsection{Generalized Evidential Deep Learning}
Although evidential deep learning has made significant progress in uncertainty modeling, it has not been utilized in the semi-supervised medical segmentation field. Here, we introduce refined uncertainty measure to generalize the evidential deep learning optimization process. Following the previous works \cite{han2021trusted,DBLP:journals/pami/HanZFZ23}, the parameters of the Dirichlet distribution are induced as:
\begin{equation}
    S^{(x,y,z)}_{v,w} = \frac{N-1}{f_{v,w}^{(x,y,z)}(C_N)},\ e^{(x,y,z)}_{v,w,n} = f^{(x,y,z)}_{v,w}(C_n) \times S^{(x,y,z)}_{v,w},\ \alpha^{(x,y,z)}_{v,w,n} = e_{v,w,n} + 1
\end{equation}

When the prediction for a voxel exhibits high uncertainty, it is preferable for the classifier to produce a negligible prediction rather than forcibly aligning it with the label, which could lead to more significant deviations in the predictions for the remaining parts. In addition, for labeled data, we utilize the student model's prediction uncertainty to guide the model to learn and correct itself from its predictions; for unlabeled data, we optimize the student network on unlabeled data using the uncertainty of the pseudo-labels generated by the teacher network. Based on this consideration, we have extended the original optimization objective of evidential deep learning by utilizing the measure of uncertainty based on the belief entropy:
\begin{equation}
    \mathcal{L}_{gedl}^r = \frac{1}{W\times H \times D}\sum_{x,y,z}(1-\overline{\mathcal{U}}(\mathcal{M}^{(x,y,z)}_{v,w}))\sum_{n=1}^{N-1}y_v^{r(x,y,z)}(log S_{v,w'}^{(x,y,z)}-\alpha^{(x,y,z)}_{v,w',n})
\end{equation}
where $r,v,w,w' = (l,i,sl,sl) \ or \ (u,j,t,su)$ and $\overline{\mathcal{U}}(\mathcal{M}^{(x,y,z)}_{v,w})$ denotes the calculated uncertainty measure after 0-1 normalization. When $r = l$, $y_v^{r(x,y,z)}$ are labels for labeled images, and it denotes pseudo-labels generated by the teacher network if $r=u$.

\subsection{Uncertainty-based Prototype Learning}
Prototypes are initially extracted from both labeled and unlabeled data independently \cite{zhang2020sg,wang2019panet}. Both types of prototypes are derived from the feature maps of the third-layer decoder, which are upscaled to match the size of the segmentation labels through trilinear interpolation. Let $h_{i,sl}^{l}$ and $h_{j,t}^{u}$ be the output feature and hidden feature of the student and teacher model for the $i_{th}$ labeled and $j_{th}$ unlabeled images. $\mathcal{B}_l$ and $\mathcal{B}_u$ represent the batch size of labeled and unlabeled set respectively and $(x, y, z)$ denotes voxel coordinate. For the labeled and unlabeled prototypes, with masked attention pooling \cite{lu2023upcol}, the uncertainty-based generation process for class $C_n$ can be unified as:
\begin{equation}
    \mathcal{P}_{C_n}^{r} = \frac{1}{\mathcal{B}_r}\sum_{v = 1}^{\mathcal{B}_r}\frac{\sum_{x,y,z}h_{v,w}^{r(x,y,z)}\beta_{v,w}^{(x,y,z)}\mathbbm{1}[y_v^{r(x,y,z)}=C_n]}{\sum_{x,y,z}\mathbbm{1}[y_v^{r(x,y,z)}=C_n]}
\end{equation}
where $r,v,w = (l,i,sl)\ or\ (u,j,t)$. We adopt the same policy as UPCoL to fuse labeled prototypes and unlabeled prototypes with dynamic proportion which derives from time-dependent Gaussian warming up function $\lambda_{con}$ \cite{tarvainen2017mean} and obtain the final labeled and unlabeled feature-to-prototype similarity $s_i^{(x,y,z)}$ and $s_j^{(x,y,z)}$ to approximate the probability of voxels in each class. Then, the labeled and unlabeled optimization objectives involving prototype consistency and generalized evidential deep learning losses can be defined as:
\begin{equation}
\begin{aligned}
    &\mathcal{L}^{l}_{total} = \sum_{x,y,z} \beta_{i,sl}^{(x,y,z)} \mathcal{L}_{CE}(s_i^{l(x,y,z)}, y_i^{l(x,y,z)}) + \mathcal{L}_{gedl}^l\\
    &\mathcal{L}^{u}_{total} = \sum_{x,y,z} \beta_{j,t}^{(x,y,z)} \mathcal{L}_{CE}(s_j^{u(x,y,z)}, y_j^{u(x,y,z)}) + \mathcal{L}_{gedl}^u
\end{aligned}
\label{eq9}
\end{equation}
where Equation \ref{eq9} represents a modification of the commonly used semi-supervised learning (SSL) technique of augmenting the training set with pseudo labels from unlabeled data. The critical distinction here is the application of a reliability-aware learning strategy on labeled and unlabeled data at the voxel level. Finally, the total loss of the proposed EPL framework is shown in Equation \ref{eq10}:
\begin{equation}
    \mathcal{L} = \mathcal{L}_{seg} + \mathcal{L}^{l}_{total} +  \lambda_{con}\mathcal{L}^{u}_{total}
    \label{eq10}
\end{equation}
where $\mathcal{L}_{seg}$ represents the average of Dice, Cross-Entropy, IoU and focal losses calculated directly with student model's prediction on labeled data and corresponding labels.

\begin{figure*}
	\centering 
	\includegraphics[scale=0.205]{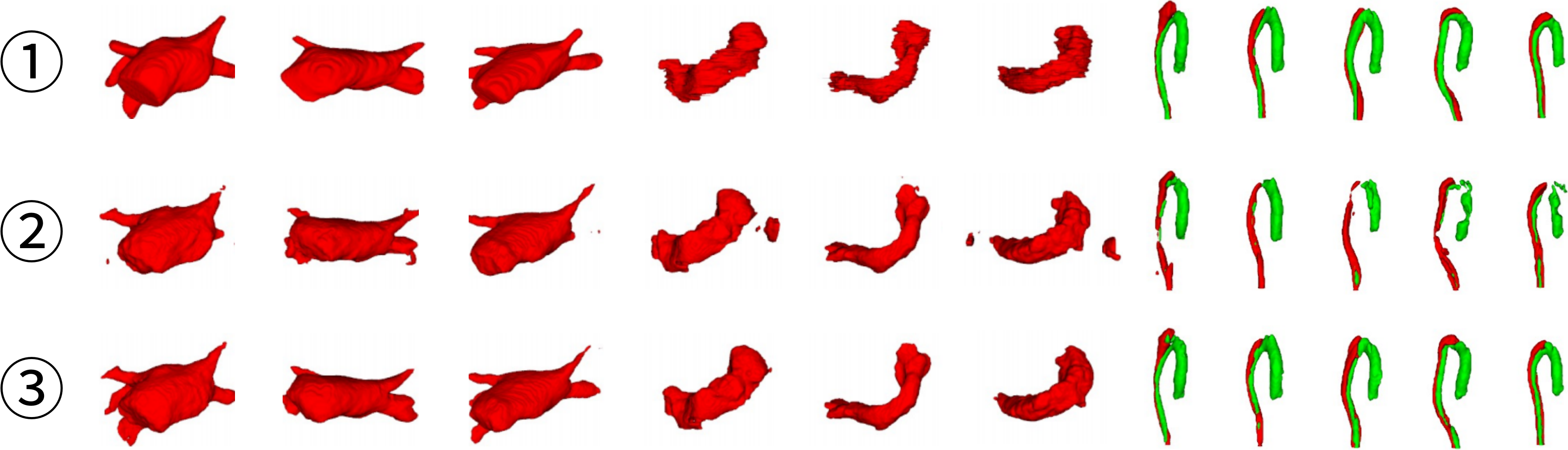}
	\caption{Comparisons of visualized results on LA, Pancreas-CT and Aortic Dissection datasets. \ding{172}, \ding{173} and \ding{174} represent the ground truth, BCP (LA, Pancreas-CT dataset) and UPCoL (TBAD dataset), the proposed method respectively. }
	\label{fig2222}
\end{figure*}
\section{Experiments}

\subsection{Datasets}
Our methodology is assessed using three distinct datasets: Pancreas-CT dataset (82 CTA scans) \cite{roth2015deeporgan}, Left Atrium dataset (LA, 100 MR images)\cite{xiong2021global}, and a multi-center type B aortic dissection (TBAD) dataset consisting of 124 CTA scans \cite{yao2021imagetbad}. The pre-processing for the Pancreas-CT and Left Atrium datasets adheres to protocols established in prior studies \cite{DBLP:conf/miccai/WuXGCZ21,xiang2022fussnet,DBLP:conf/miccai/YuWLFH19}. The TBAD dataset, meticulously labeled by seasoned radiologists, is divided into 100 training scans and 24 test scans, incorporating both publicly available data and data gathered by UPCoL. This dataset is standardized to a resolution of 1mm$^3$ and adjusted to dimensions of 128×128×128, following established guidelines \cite{cao2019fully}. For each of these datasets, we utilize merely 20\% of the labeled training data at most, normalizing the voxel intensities to have a zero mean and unit variance.
\begin{table*}[htbp]\scriptsize
        \centering
	\renewcommand{\arraystretch}{1.4}
        \caption{The performance of V-Net on Left Atrium and Pancreas-CT dataset in labeled ratio 5$\%$, 10$\%$, 20$\%$ and 100$\%$}
		\setlength{\tabcolsep}{1.13mm}{
			\begin{tabular}{c |c c |c c c c| c c c c c c c c}
				\hline
                    Dataset& \multicolumn{2}{c|}{\multirow{2}*{Scans Used}} &\multicolumn{4}{c|}{LA dataset}& \multicolumn{4}{c}{Pancreas-CT dataset}\\\cline{4-11}\cline{1-1}
				\multicolumn{1}{c|}{\multirow{2}*{Model}}&{} &{} &\multicolumn{4}{c|}{Metrics}&\multicolumn{4}{c}{Metrics}  \\\cline{2-11} 
                &{Labeled} & {Unlabled} & Dice$\uparrow$ & Jaccard$\uparrow$ & 95HD$\downarrow$ & ASD$\downarrow$& Dice$\uparrow$ & Jaccard$\uparrow$ & 95HD$\downarrow$ & ASD$\downarrow$\\\cline{1-11} 
                \multirow{4}*{V-Net}&(5\%)&(95\%)& 52.55 & 39.60 & 47.05 & 9.87&55.06&40.48&32.86&12.67\\
                &(10\%)&(90\%)& 82.74 & 71.72 & 13.35 & 3.26&69.65&55.19&20.24&6.31\\
                &(20\%)&(80\%)& 87.84 & 78.56 & 9.10 & 2.65 & 71.27 & 55.71 &17.48   & 5.16\\
                &(100\%)&(0\%)& \textbf{92.62} & \textbf{85.24} & \textbf{4.47} & \textbf{1.33}&\textbf{85.74}&\textbf{73.86}&\textbf{4.48}&\textbf{1.07}\\\hline
		\end{tabular}}
 \label{1}
\end{table*}

\begin{table*}[htbp]\scriptsize
	\label{table5}
	\centering
        \renewcommand{\arraystretch}{1.3}
        \caption{Comparisons with state-of-the-art models on Left Atrium dataset in labeled ratio 5$\%$, 10$\%$ and 20$\%$}
			\setlength{\tabcolsep}{0.7mm}{\begin{tabular}{c | c | c c c c c c c c c c c c}
                \hline
                Labeled Ratio & Metrics  & UA-MT & SASSNet & DTC & URPC & MC-Net & SS-Net & Co-BioNet & BCP & Ours\\
                \hline
                \multirow{4}*{(5\%)}   & Dice$\uparrow$   &82.26 & 81.60 & 81.25& 86.92&87.62&  86.33 &76.88 &88.02 & \textcolor{red}{\textbf{90.85}}\\
                & Jaccard$\uparrow$ &70.98 & 69.63 & 69.33 & 77.03&78.25 &76.15 &66.76 &78.72&\textcolor{red}{\textbf{83.32}} \\
                & 95HD$\downarrow$ & 13.71&16.16 &14.90   &  11.13&  10.03&9.97& 19.09 &7.90&\textcolor{red}{\textbf{6.08}}  \\
                & ASD$\downarrow$ & 3.82&3.58&3.99& 2.28& 1.82& 2.31&  2.30 & 2.15 & \textcolor{red}{\textbf{1.76}}\\ \hline
                \multirow{4}*{(10\%)}   & Dice$\uparrow$  & 87.79 & 87.54 &87.51&86.92& 87.62 & 88.55 &  89.20& 89.62 & \textcolor{red}{\textbf{91.80}}\\
                & Jaccard$\uparrow$ & 78.39 &  78.05 & 78.17& 77.03& 78.25 & 79.62& 80.68& 81.31 & \textcolor{red}{\textbf{84.91}}\\
                & 95HD$\downarrow$ & 8.68 & 9.84&  8.23&11.13& 10.03 & 7.49&  6.44&  6.81 & \textcolor{red}{\textbf{5.07}}\\
                & ASD$\downarrow$ &  2.12 &  2.59& 2.36& 2.28 &  1.82 & 1.90&  1.90& 1.76 & \textcolor{red}{\textbf{1.54}}\\ \hline
                \multirow{4}*{(20\%)}   & Dice$\uparrow$ & 88.88& 89.54& 89.42& 88.43& 90.12&  89.25&  91.26 &90.34 & \textcolor{red}{\textbf{92.30}}\\
                & Jaccard$\uparrow$ &80.21& 81.24&80.98 &81.15&82.12&81.62& 83.99&  82.50 & \textcolor{red}{\textbf{85.72}}\\
                & 95HD$\downarrow$ & 7.32& 8.24& 7.32& 8.21& 11.28&  6.45&  5.17& 6.75 & \textcolor{red}{\textbf{4.73}}\\
                & ASD$\downarrow$ &  2.26&1.99& 2.10&2.35&2.30& 1.80&  1.64&  1.77 & \textcolor{red}{\textbf{1.38}}\\ \hline
		\end{tabular}}
 \label{2}
\end{table*}

\subsection{Implementation Details}
We utilize V-Net \cite{DBLP:conf/3dim/MilletariNA16} as our foundational architecture, benchmarking the performance of V-Nets trained with 20\% and 100\% labeled data as the minimal and maximal performance thresholds, respectively. Within the mean-teacher framework, the student network undergoes training for 10,000 iterations utilizing the Adam optimizer with a learning rate 0.001. Concurrently, the teacher network's parameters are refined through the exponential moving average (EMA) based on the student's parameters. The training batch size is 3, comprising one labeled and two unlabeled samples. Consistent with prior studies \cite{xiang2022fussnet,lu2023upcol}, we employ random cropping to generate cubes measuring 96$\times$96$\times$96 and 112$\times$112$\times$80 for the Pancreas-CT and Left Atrium datasets, respectively. Moreover, we implement 3-fold cross-validation and enhance the training process with data augmentation techniques such as rotation (within a range of -10$^{\circ}$ to 10$^{\circ}$) and scaling (zoom factors ranging from 0.9 to 1.1) for the TBAD dataset, following established protocols \cite{cao2019fully,fantazzini20203d}. To evaluate performance, we employ four metrics: Dice coefficient (Dice), Jaccard Index (Jaccard), 95\% Hausdorff Distance (95HD), and Average Symmetric Surface Distance (ASD).

\begin{table*}[htbp] \scriptsize
	\label{table5}
	\centering
        \renewcommand{\arraystretch}{1.4}
        \caption{Comparisons with state-of-the-art models on Pancreas-CT dataset in labeled ratio 5$\%$, 10$\%$ and 20$\%$}
			\setlength{\tabcolsep}{0.7mm}{\begin{tabular}{c | c | c c c c c c c c c c c c}
                \hline
                Labeled Ratio & Metrics  & UA-MT & SASSNet & DTC & URPC & MC-Net & SS-Net & Co-BioNet & BCP & Ours\\
                \hline
                \multirow{4}*{(5\%)}   & Dice$\uparrow$ &  47.03&56.05&49.83&52.05& 54.99&56.35& 79.74& 80.33&\textcolor{red}{\textbf{82.83}}\\
                & Jaccard$\uparrow$ &  32.79& 41.56& 34.47&36.47&40.65&43.41& 65.66& 67.65& \textcolor{red}{\textbf{71.04}}\\
                & 95HD$\downarrow$ & 35.31&36.61& 41.16&34.02&16.03& 22.75&5.43& 11.78 & \textcolor{red}{\textbf{4.20}}\\
                & ASD$\downarrow$ &  4.26&4.90& 16.53&13.16& 3.87& 5.39&2.79&4.32 & \textcolor{red}{\textbf{1.24}}\\  \hline
                \multirow{4}*{(10\%)}   & Dice$\uparrow$ &66.96&66.69& 67.28& 64.73&69.07&67.40& 82.49& 81.54&\textcolor{red}{\textbf{83.78}}\\
                & Jaccard$\uparrow$ &  51.89& 51.66& 52.86&49.62& 54.36&53.06& 67.88&69.29&\textcolor{red}{\textbf{72.32}}\\
                & 95HD$\downarrow$ & 21.65 & 18.88& 17.74&21.90& 14.53&20.15& 6.51&12.21&\textcolor{red}{\textbf{4.36}}\\
                & ASD$\downarrow$ &  6.25&5.76& 1.97& 7.73& 2.28& 3.47& 3.26& 3.80&\textcolor{red}{\textbf{1.44}}\\ \hline
                \multirow{4}*{(20\%)}   & Dice$\uparrow$ & 77.26 &77.66&78.27& 79.09& 78.17& 79.74&84.01&82.91 & \textcolor{red}{\textbf{84.63}}\\
                & Jaccard$\uparrow$ &  63.82&64.08& 64.75&65.99& 65.22& 65.42&70.00&70.97 & \textcolor{red}{\textbf{74.56}} \\
                & 95HD$\downarrow$ &  11.90&10.93& 8.36& 11.68& 6.90& 12.44&\textcolor{red}{\textbf{5.35}}&6.43& 5.86\\
                & ASD$\downarrow$ & 3.06& 3.05& 2.25& 3.31&\textcolor{red}{\textbf{1.55}}& 2.69&2.75& 2.25& 2.42\\ \hline
		\end{tabular}}
 \label{3}
\end{table*}
\subsection{Results on the Left Atrium Dataset and Pancreas-CT Dataset}
For the Left Atrium dataset and Pancreas-CT dataset, the method proposed in this paper is able to achieve performance very close to that of the fully supervised V-Net with only 20\% of the labeled data utilized and realize a reversal in the ASD metric, which is presented in Table \ref{1}. Besides, the proposed method is also compared with other eight state-of-the-art semi-supervised medical image segmentation methods, such as
UA-MT \cite{DBLP:conf/miccai/YuWLFH19}, SASSNet \cite{DBLP:conf/miccai/LiZH20}, DTC \cite{DBLP:conf/aaai/LuoCSW21}, URPC \cite{DBLP:conf/miccai/LuoLCSCZCWZ21}, MC-Net \cite{DBLP:conf/miccai/WuXGCZ21}, SS-Net \cite{DBLP:conf/miccai/WuWWGC22}, Co-BioNet \cite{peiris2023uncertainty} and BCP \cite{DBLP:conf/cvpr/BaiCL0023}. The detailed results are provided in Table \ref{2} and \ref{3}. In summary, in the experiments based on the LA and Pancreas-CT datasets, the method proposed in this paper achieves almost the best segmentation performance across three different labeled ratios. Moreover, it realizes considerable improvements over the two previously best methods, BCP and CoBioNet. For instance, in the 20\% labeled ratio for the Dice and Jaccard scores, the proposed method achieves improvements of more than 1\% on both datasets. Among them, the most significant improvements are observed in the Jaccard metrics, which achieve considerable progress compared to the previous methods and further illustrate the superiority of the proposed method. Visualization results for the Left Atrium dataset and Pancreas-CT datasets are provided in Fig. \ref{fig2222}. Additionally, a considerable portion of the segmentation results from the comparative methods is fragmented and not connected to the parts that should be segmented, which is completely inconsistent with reality. In summary, it can be observed that the segmentation results of the method proposed in this paper are more consistent with the ground truth compared to comparative methods. These experimental results strongly demonstrate the effectiveness of the evidential fusion-based multi-classifier predictive fusion and the proposed uncertainty measurement in guiding model learning.

\begin{table*}[htbp] \tiny
	\renewcommand{\arraystretch}{1.4}
	\label{table5}
	\centering
        \caption{Comparisons with state-of-the-art models on Aortic Dissection dataset in labeled ratio 5$\%$, 10$\%$ and 20$\%$}
		\setlength{\tabcolsep}{1mm}{
			\begin{tabular}{c |c c |c c c |c c c| c c c |c c c  }
				\hline
                    \multicolumn{1}{c|}{\multirow{3}*{Model}}& \multicolumn{2}{c|}{\multirow{2}*{Scans Used}} &\multicolumn{12}{c}{Metrics}\\\cline{4-15}
				{}&{} &{} &\multicolumn{3}{c|}{Dice$\uparrow$}&\multicolumn{3}{c|}{Jaccard$\uparrow$}&\multicolumn{3}{c|}{95HD$\downarrow$}&\multicolumn{3}{c}{ASD$\downarrow$}  \\\cline{2-15}
                &{Labeled} & {Unlabled} & TL & FL & Mean & TL & FL & Mean & TL & FL & Mean& TL & FL & Mean \\ \cline{1-15}
                \multirow{2}*{V-Net}&(20\%)&(80\%)&55.51& 48.98 &52.25&39.81 &34.79& 37.30 &7.24& 10.17& 8.71& 1.27 &3.19 &2.23 \\
                &(100\%)&(0\%)&  75.98& 64.02& 70.00& 61.89& 50.05& 55.97& 3.16& 7.56& 5.36 &0.48& 2.44 &1.46 \\\hline
				MT&\multirow{5}*{(20\%)} & \multirow{5}*{(80\%)} &  57.62& 49.95& 53.78& 41.57& 35.52&38.54& 6.00& 8.98& 7.49& 0.97& 2.77& 1.87 \\
				UA-MT&&&70.91& 60.66 &65.78& 56.15& 46.24& 51.20& 4.44& 7.94& 6.19& 0.83& 2.37& 1.60  \\
				FUSSNet&&&  79.73& 65.32& 72.53& 67.31& 51.74& 59.52& 3.46& 7.87& 5.67& 0.61& 2.93& 1.77  \\
				URPC&&&  81.84& 69.15& 75.50& 70.35& 57.00& 63.68& 4.41& 9.13& 6.77& 0.93& \textcolor{red}{\textbf{1.11}} &1.02  \\
				UPCoL&&&  82.65& 69.74& 76.19& 71.49& 57.42& 64.45& 2.82& 6.81& 4.82& 0.43& 2.22& 1.33 \\ \hline
				\multirow{3}*{Ours}&(5\%)&(95\%)&\textcolor{red}{\textbf{82.72}}& \textcolor{red}{\textbf{74.82}} & \textcolor{red}{\textbf{78.77}} & 71.12&\textcolor{red}{\textbf{62.09}}&\textcolor{red}{\textbf{66.61}}&\textcolor{red}{\textbf{2.54}}&\textcolor{red}{\textbf{4.66}}&\textcolor{red}{\textbf{3.61}} & \textcolor{red}{\textbf{0.32}} & 1.35 & \textcolor{red}{\textbf{0.98}}\\&(10\%)&(90\%)&\textcolor{red}{\textbf{82.87}}& \textcolor{red}{\textbf{75.79}} & \textcolor{red}{\textbf{79.33}} & 71.22 &\textcolor{red}{\textbf{63.26}}&\textcolor{red}{\textbf{67.24}}&\textcolor{red}{\textbf{2.51}}&\textcolor{red}{\textbf{4.85}}&\textcolor{red}{\textbf{3.68}} & \textcolor{red}{\textbf{0.43}} & 1.42 & \textcolor{red}{\textbf{0.92}}\\
                &(20\%)&(80\%)&\textcolor{red}{\textbf{86.14}}& \textcolor{red}{\textbf{77.32}} & \textcolor{red}{\textbf{81.73}} & \textcolor{red}{\textbf{76.26}}&\textcolor{red}{\textbf{65.33}}&\textcolor{red}{\textbf{70.80}}&\textcolor{red}{\textbf{2.02}}&\textcolor{red}{\textbf{4.76}}&\textcolor{red}{\textbf{3.39}} & \textcolor{red}{\textbf{0.34}} & 1.32 & \textcolor{red}{\textbf{0.83}}\\
                \hline
		\end{tabular}}
 \label{4}
\end{table*}

\subsection{Results on the Aortic Dissection Dataset}
For Aortic Dissection dataset, the proposed method is compared with other five state-of-the-art semi-supervised medical image segmentation methods, such as MT \cite{tarvainen2017mean}, UA-MT \cite{DBLP:conf/miccai/YuWLFH19}, FUSSNet \cite{xiang2022fussnet}, URPC \cite{DBLP:conf/miccai/LuoLCSCZCWZ21} and UPCoL \cite{lu2023upcol}. The detailed results are provided in Table \ref{4}. It can be deduced from the table that the method proposed in this paper is able to achieve nearly an 7\% performance advantage over the fully supervised V-Net with only 5\% of the labeled data, and with 20\% of the labeled data utilized, it can achieve over an 10\% performance gain compared to the fully supervised V-Net. Compared to other semi-supervised methods, for instance, the previous state-of-the-art (SOTA) method UPCoL, the method proposed in this paper still manages to surpass its performance with 20\% labeled data while using only 5\% of the labeled data. The visualized results of the Aortic Dissection dataset are provided in Fig. \ref{fig2222}. The segmentation results of the comparative methods are not coherent enough, particularly in the green areas of the visualized results, where they perform poorly. The proposed method, on the contrary, maintains a very high similarity in the segmentation results of the green areas compared to the ground truth. It can be concluded that the output of the proposed method is much better than those of UPCoL, which achieves results very close to the ground truth. Overall, on the TBAD dataset, the method proposed in this paper realize significant performance gains, which once again demonstrates the effectiveness of the generalized evidential deep learning and the proposed uncertainty measurement method.

\subsection{Ablation Study}
In this study, we utilize the Pancreas-CT dataset with 20\% labeled data to explore the role of each key component designed in our method, including MT (mean-teacher architecture), AMC (average multi-classifier), PL (prototype learning with fusion), EFMC (evidential fusion-based multi-classifier), LRM (labeled reliability map), URM (unlabeled reliability map), LEDL (labeled evidential deep learning), and UEDL (unlabeled evidential deep learning). The detailed results of the ablation study are presented in Table \ref{ablation}.

\begin{table*}[h]\scriptsize
\renewcommand{\arraystretch}{1.4}
\caption{Experimental results of ablation study on Pancreas-CT dataset}
    \centering
\setlength{\tabcolsep}{1mm}{
    \begin{tabular}{c|cccccccc|cccccc}
        \hline
        & MT & AMC & PL & EFMC & LRM & URM & LEDL & UEDL & Dice$\uparrow$ & Jaccard$\uparrow$ & 95HD$\downarrow$ & ASD$\downarrow$\\\hline
        MT& \Checkmark&&&&&&&&76.04&61.92 & 20.56 & 6.18    \\
        +AMC& \Checkmark & \Checkmark&&&&&&&79.91&67.03 & 6.37 & 1.68    \\
        +PL& \Checkmark & \Checkmark &\Checkmark&&&&&& 80.99&68.44 & 6.76&2.33    \\
        +EFMC& \Checkmark & \Checkmark & \Checkmark &\Checkmark&&&&&81.68&69.31&6.02&2.19  \\
        + LRM& \Checkmark & \Checkmark &\Checkmark& \Checkmark&\Checkmark&&&&82.76&70.83& 5.48&1.99   \\
        + URM& \Checkmark & \Checkmark & \Checkmark & \Checkmark &\Checkmark & \Checkmark&&&83.74&72.32&\textbf{4.38}&\textbf{1.64} \\
        + LEDL& \Checkmark & \Checkmark & \Checkmark& \Checkmark & \Checkmark & \Checkmark & \Checkmark &&84.15&74.01&6.81&1.98 \\
        EPL& \Checkmark & \Checkmark & \Checkmark& \Checkmark & \Checkmark & \Checkmark & \Checkmark &\Checkmark&\textbf{84.63}&\textbf{74.56}&5.86&2.42 \\
        \hline
        
    \end{tabular}}
    \label{ablation}
\end{table*} 

Initially, we conduct experiments on a network structure with only the MT component, noting that the MT structure incurs a cost to maintain consistency between the student and teacher networks. It is found that under such a configuration, the MT-structured network surpassed the network with only V-Net, achieving approximately a 5\% improvement on the Dice metric, with various degrees of performance fluctuations on other three metrics. When structures of multiple classifiers are added to the model, the model's performance further improves, indicating that multiple classifiers could help the network find more suitable segmentation results. We incorporate the prototype fusion mechanism from UPCoL without introducing the corresponding reliability map and observe that this mechanism could enhance model performance after the introduction of fused prototypes. Thus far, we have explored the basic structure of the proposed method on performance. Next, we further analyze the roles of several designs proposed in enhancing the model's segmentation capabilities. The introduction of EFMC significantly enhanced the performance of the AMC design, likely due to the fusion rules considering the interrelations and uncertainties of each category during synthesis and measurement. Subsequently, we introduce LRM and URM for measuring uncertainties in labeled and unlabeled learning parts, respectively, and apply them in the prototype generation process. Table data analysis reveals that the reliability map generated for labeled learning improves the overall segmentation performance of the model, though the improvement on the 95HD metric is not as pronounced as when EFMC was introduced. With respect to the reliability map, URM contributes more to the model performance improvement than LRM across the four measure metrics, possibly because uncertainty of the pseudo-labels generated in the unlabeled part have more influence on the segmentation of objects. It can be concluded that URM plays a crucial role in effectively screening out unreliable features during the unlabeled prototype generation process. Lastly, we discuss the generalized evidential deep learning objective for both labeled and unlabeled parts. The introduction of generalized evidential deep learning in the labeled and unlabeled parts further improves model performance, as there is considerable uncertainty in the two kinds of learning process. However, it is worth noting that generalized evidential deep learning is not very friendly to the index measuring boundary overlap when the proportion of labeled data is high. Overall, the components of the generalized evidential deep learning framework greatly assist in enhancing model performance, especially the uncertainty-guided learning process for unlabeled data, which significantly improves the model segmentation capability.


\section{Conclusion}
The evidential prototype learning framework advances semi-supervised learning by enhancing prediction accuracy through refined uncertainty handling. It broadens the probabilistic model, employs Dempster’s combination rule for better prediction fusion, and introduces a dual uncertainty measurement combining evidence uncertainty and belief entropy. This allows for a self-correcting learning process and more strategic guidance on unlabeled data. The redesigned optimization function minimizes the influence of highly uncertain parts, maintaining consistency with labeled data. The approach significantly improves model performance in uncertain environments by mitigating the impact of unreliable features, marking a robust step forward in leveraging uncertainty for more accurate and reliable predictions in semi-supervised learning scenarios. Extensive experiments demonstrate that the proposed EPL framework is competitive with existing approaches on Left Atrium, Pancreas-CT and TBAD datasets.

\bibliographystyle{splncs04}
\bibliography{ref}
\end{document}